\documentclass{article}
\usepackage{ijcai19}

\usepackage{times}
\usepackage{soul}
\usepackage{url}
\usepackage[hidelinks]{hyperref}
\usepackage[utf8]{inputenc}
\usepackage[small]{caption}
\usepackage{graphicx}
\usepackage{amsmath}
\usepackage{booktabs}
\usepackage{algorithm}
\usepackage{algorithmic}
\usepackage{epsfig}
\usepackage{amssymb}
\usepackage{multirow}
\usepackage{array}
\usepackage{color}

\title{Transferable Adversarial Attacks for Image and Video Object Detection}

\author{Xingxing Wei$^1$, Siyuan Liang$^2$, Ning Chen$^1$, Xiaochun Cao$^2$\\
$^1$Department of Computer Science and Technology, Tsinghua University\\
$^2$Institute of Information Engineering, Chinese Academy of Sciences\\
\{xwei11, ningchen\}@mail.tsinghua.edu.cn, \{liangsiyuan, caoxiaochun\}@iie.ac.cn
}
\begin{document}

\maketitle

\begin{abstract}
Identifying adversarial examples is beneficial for understanding deep networks and developing robust models. 
However, existing attacking methods for image object detection have two limitations: {\it weak transferability}---the generated adversarial examples often have a low success rate to attack other kinds of detection methods, and {\it high computation cost}---they need much time to deal with video data, where many frames need polluting. To address these issues, we present a generative method to obtain adversarial images and videos, thereby significantly reducing the processing time. To enhance transferability, we manipulate the feature maps extracted by a feature network, which usually constitutes the basis of object detectors. Our method is based on the Generative Adversarial Network (GAN) framework, where we combine a high-level class loss and a low-level feature loss to jointly train the adversarial example generator. Experimental results on PASCAL VOC and ImageNet VID datasets show that our method efficiently generates image and video adversarial examples, and more importantly, these adversarial examples have better transferability, therefore being able to simultaneously attack two kinds of  representative object detection models: proposal based models like Faster-RCNN and regression based models like SSD. 
\end{abstract}

\begin{figure}[t]
\centering\includegraphics[width=0.43\textwidth]{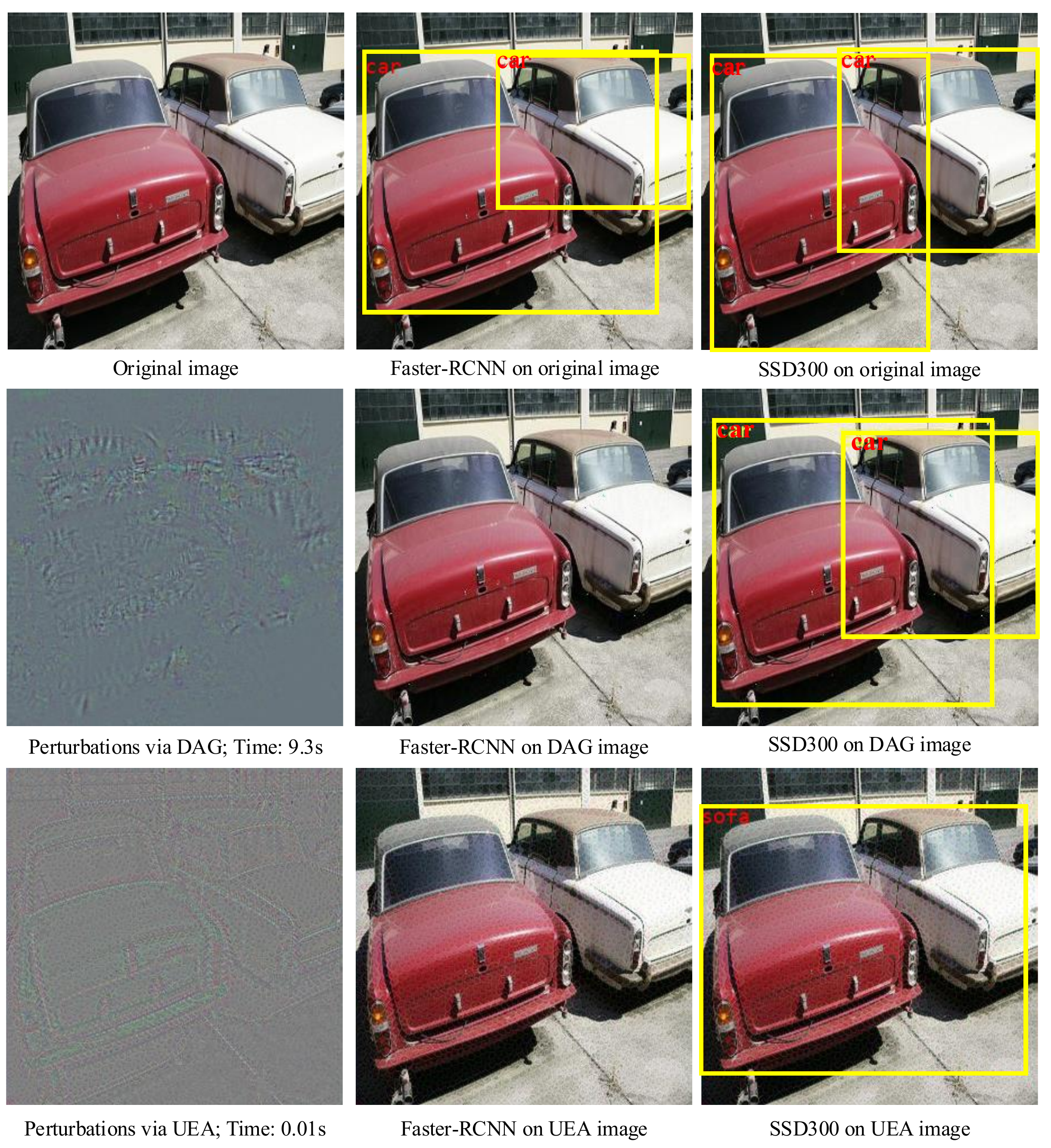}
\caption{An example of the comparisons  between DAG (Dense Adversary Generation) and our UEA (Unified and Efficient Adversary) against proposal and regression based detectors. In the first row, Faster-RCNN and SSD300 detect the correct objects. The second row lists the adversarial examples from DAG. We see it succeeds to attack Faster-RCNN, but fails to attack SSD300. In this third row, neither Faster-RCNN nor SSD300 detects the cars on the adversarial images. Moreover, the UEA's processing time  is almost 1000 times faster than DAG for generating an adversarial image. }
\label{fig:figure1}
\vspace{-.4cm}
\end{figure}

\section{Introduction}

Deep learning techniques have achieved great success in various computer vision tasks \cite{li2018textbook,wei2018video}. However, it is also proved that neural networks are vulnerable to adversarial examples \cite{szegedy2013intriguing}, thereby attracting a lot of attention on attacking (e.g., FGSM \cite{goodfellow2014explaining,Dong_2018_CVPR}, deepfool \cite{moosavi2016deepfool}, C\&W attack \cite{carlini2017towards}) and defending~(e.g.,~\cite{raghunathan2018certified}) a network. Attacking is beneficial for deeply understanding neural networks~\cite{dong2017towards} and motivating more robust solutions~\cite{pang2018towards}. 
Though much work has been done for image classification, more and more methods are presented to attack other tasks~\cite{akhtar2018threat}, such as face recognition~\cite{sharif2016accessorize},
video action recognition~\cite{wei2018sparse}, and the physical-world adversarial attack on road signs~\cite{evtimov2017robust}.

\iffalse
For example, \cite{sharif2016accessorize} designs the real and stealthy attacks on state-of-the-art face recognition. \cite{wei2018sparse} proposes the sparse adversarial perturbations for video action recognition.  \cite{evtimov2017robust} explores the physical-world attack for road signs detection.  We refer readers to  \cite{akhtar2018threat} for a survey of adversarial attacks on computer vision.
\fi

\begin{figure*}[t]
\centering\includegraphics[width=0.93\textwidth]{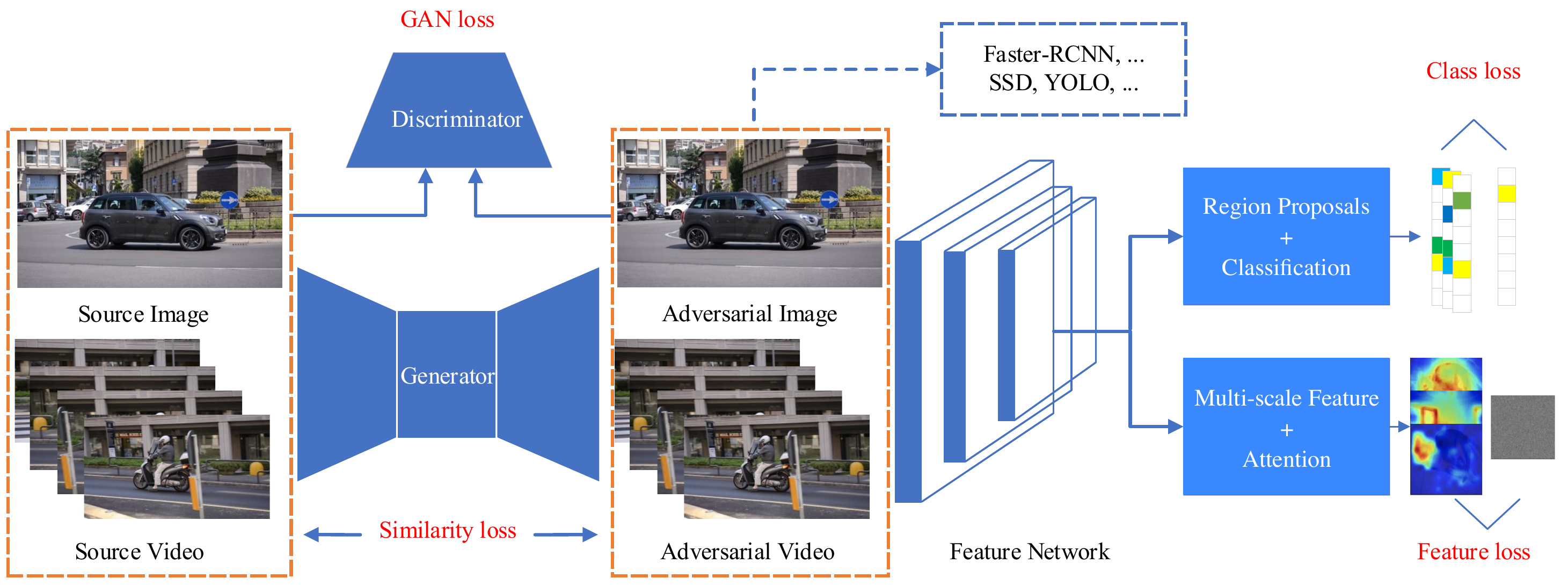}
\caption{The training framework of Unified and Efficient Adversary (UEA). Besides the GAN loss and similarity loss, we formulate DAG's high-level class loss and our low-level \textbf{multi-scale attention feature loss} into GAN framework to jointly train a generator. In the testing phase,  the generator is used to output adversarial images or video frames to  fool the different classes of object detectors (blue dashed box). }
\label{fig:figure2}
\vspace{-.4cm}
\end{figure*}

As the core task in computer vision, object detection for image data has also been attacked. It is known that the current object detection models can be roughly categorized into two classes: proposal based models and regression based models. The various mechanisms make attacking object detection more complex than image classification.
\cite{xie2017adversarial} proposes a white-box attacking method for proposal based models: Dense Adversary Generation (DAG). They choose Faster-RCNN \cite{ren2017faster} as the threat model. DAG firstly assigns an adversarial label for each proposal region and then performs iterative gradient back-propagation to misclassify the proposals. The similar methods are also presented  in \cite{chen2018robust,li2018robust}. 
Because regression-based methods don't use region proposals, DAG cannot directly transfer to attack them. That means DAG has weak black-box attacking ability.
In addition, DAG is an optimization method, which often needs 150 to 200 iterations to meet the end for an image~\cite{xie2017adversarial}. The high computation cost makes DAG not available for attacking video object detection, 
which usually considers temporal interactions between  adjacent frames \cite{zhu2017flow,zhu2017deep} and therefore the most reliable attacking method for video object detection is to pollute  all the  frames or many key frames in the video. 

To address these issues, in this paper, we propose the Unified and Efficient Adversary (UEA) for image and video object detection. ``Efficient" specifies that our method is able to quickly generate adversarial  images, and thus can efficiently deal with every frame in the video. To this end,  we utilize a generative mechanism instead of the optimization procedure. Specifically, we formulate the problem into Generative Adversarial Network (GAN) framework like \cite{xiao2018generating,Poursaeed_2018_CVPR}, and train a generator network to generate adversarial images and key frames. Because the testing step only involves the forward network, the running time is fast. As for ``Unified", it means that the proposed adversary has better transferability than DAG, and thus has strong black-box attacking ability.  It can not only perform reliable attack to Faster-RCNN like DAG, but also effectively attack regression based detectors. We observe that both the proposal and regression based detectors utilize feature networks as their backends. For examples, Faster-RCNN and SSD \cite{liu2016ssd} use the same VGG16 \cite{simonyan2014very}. If we manipulate the features maps in Faster-RCNN, the generated adversarial examples will also make SSD fail to detect objects. This idea is implemented as a multi-scale attention feature loss in our paper, i.e., manipulating the feature maps from multiple layers. To fool detectors, only the regions of foreground objects need perturbing. Therefore, an attention weight is integrated into the feature loss to manipulate the feature subregions.  The usage of attention weight also improves the imperceptibility of generated adversarial examples because the number of perturbed pixels is limited. In the viewpoint of DNNs' depth, DAG's class loss is applied on the high-level softmax layer,  and attention feature loss is performed on the low-level backend layer. Besides class loss, UEA incorporates an additional feature loss to get the strong transferability, which is reasonable. 
Figure \ref{fig:figure1} gives an example of UEA, and Figure \ref{fig:figure2} illustrates the overall framework.

In summary, this paper has the following contributions:
\begin{itemize}
\item We propose the Unified and Efficient Adversary (UEA) for attacking image and video detection. To the best of our knowledge, UEA is the first attacking method that can not only efficiently deal with both images and videos, but also simultaneously fool the proposal based detectors and regression based detectors. 

\item We propose a multi-scale attention feature loss to enhance the UEA's black-box attacking ability. Furthermore, we formulate the existing high-level class loss and the proposed low-level feature loss within GAN framework to jointly train a better generator.  
\end{itemize}

The rest of this paper is organized as follows. In Section 2, we briefly review the related work. We present the proposed Unified and Efficient Adversary framework in Section 3. Section 4 reports all experimental results. Finally, we summarize the conclusion in Section 5.

\section{Related Work}
The related work comes from two aspects: image and video object detection and adversarial attack for object detection. 

\subsection{Image and Video Object Detection}
Object detection is an area where deep learning has shown its great power. Currently, the dominant image object detection models can be roughly categorized into two classes: proposal based models and regression based models. The former class typically contains R-CNN \cite{girshick2016region},  Faster-RCNN \cite{ren2017faster}, Mask-RCNN \cite{he2017mask}, etc. These kinds of methods use a two-step procedure. They firstly detect proposal regions, and then classify them to output the final detected results. The latter class is represented by YOLO \cite{redmon2016you} and SSD \cite{liu2016ssd}. They regard the detection task as the regression process, and directly predict the coordinates of bounding boxes.  Compared with the image scenario, video object detection incorporates temporal  interactions between adjacent frames into the procedure. They usually apply the existing image detector on the selected key frames, and then propagate the bounding boxes via temporal interactions \cite{zhu2017flow,zhu2017deep,chen2018optimizing}. Therefore, image object detection forms the basis of the video object detection. In this paper, we aim to present a unified method that can attack both the image and video detectors.

\subsection{Adversarial Attack for Object Detection}
Currently, adversarial attacks for the object detection are rare. The first method is proposed by \cite{xie2017adversarial}, named DAG. They firstly assign an adversarial label for each proposal region and then perform iterative gradient back-propagation to misclassify the proposals. DAG is based on the optimization, and is time consuming, it needs many iterations to accomplish an adversarial image.  \cite{chen2018robust,li2018robust}  present the similar idea. In addition,  \cite{bose2018adversarial} tries to attack the face detector. But their threat model is also based on proposal based detectors (Faster-RCNN).
All these works attack the proposal based object detectors, and they are all based on the the optimization manner. A unified adversary, which can simultaneously attack both the proposal based and regression based detectors, is absent. In this paper, we aim to fill in this gap, and present a unified method that can attack both the detectors. 

\begin{table}[htbp]
\vspace{-.1cm}
\scriptsize
\caption{The difference between our method and the existing attacking methods for object detection. } \centering
\vspace{-.2cm}
\begin{tabular}{|c|c|c|c|c|}
\hline
\textbf{Methods}          &Image  &Video &Proposal &Regression    \\
\hline
\cite{bose2018adversarial}  &\checkmark  & &\checkmark &  \\
\hline
\cite{chen2018robust}   &\checkmark  & &\checkmark &  \\
\hline
\cite{li2018robust}   &\checkmark  & &\checkmark &  \\
\hline
\cite{xie2017adversarial}  &\checkmark  &  &\checkmark &  \\
\hline
Our UEA    &\checkmark  &\checkmark &\checkmark  &\checkmark  \\
\hline
\end{tabular}\label{tab:tab1}
\vspace{-.6cm}
\end{table}

\section{Methodology}
In this section, we introduce the details of UEA. 
\subsection{Problem Definition}
Given an image $I$, our goal is to generate its corresponding adversarial image $\hat{I}$. We hope that $\hat{I}$ can attack the object detector $Dt$. For a ground-truth object $(B_i, C_i)$ on $I$, where $B_i$ is the bounding box, and $C_i$ is the label. Suppose the object detector $Dt$ succeeds to detect this object and outputs $(b_i, c_i)$, where the IOU between $B_i$ and $b_i$ is more than 0.5, and $C_i=c_i$. We let $(\hat{b}_i,\hat{c}_i)$ denote the detected result of this object on the adversarial image $\hat{I}$ (Note that $\hat{b}_i$ may be empty, which represents $Dt$ doesn't detect this object). If the IOU between $\hat{b}_i$ and $B_i$ is less than 0.5 or $\hat{c}_i\neq C_i$, we can say the object detector $Dt$ is successfully attacked or fooled. In order to measure the performance of attacking methods, we will compute the detection accuracy using mAP (mean Average Precision) on the entire dataset, and check the mAP drop after attacks. For videos, we regard the key frames in a video as images, and perform the same operation. We expect the adversarial video can also fool the state-of-the-art video detection models. The $Dt$ is based on proposals or regression. 

\subsection{Unified and Efficient Adversary}
 In this section, we introduce the technical details of UEA. Overall, we utilize a generative mechanism to accomplish this task. Specifically, we formulate our problem into the conditional GAN framework. The objective of the conditional GAN can be expressed as:
%----------------------------Eq.1------------------------------------
\begin{equation}\label{eq:eq1}
\mathcal{L}_{cGAN}(\mathcal{G,D})= \mathbb{E}_I[\log\mathcal{D}(I)]+\mathbb{E}_I[\log(1-\mathcal{D}(\mathcal{G}(I)))],
\end{equation}
%--------------------------------------------------------------------
where $\mathcal{G}$ is the generator to compute adversarial examples, and $\mathcal{D}$ is the discriminator to distinguish the adversarial examples from the clean images.  Because adversarial examples are defined as close as  as possible with original examples \cite{szegedy2013intriguing}, we input the original images (or frames) and adversarial images (or frames) to the discriminator to compute GAN loss in Eq.(\ref{eq:eq1}). In addition, an $L_2$ loss between the clean images (or frames) and adversarial images (or frames) is applied to measure their similarity:
%----------------------------Eq.2------------------------------------
\begin{equation}\label{eq:eq2}
\mathcal{L}_{L_2}(\mathcal{G})= \mathbb{E}_I[\vert\vert I-\mathcal{G}(I)\vert\vert_2].
\end{equation}
%--------------------------------------------------------------------
After training the generator based on GAN framework, we use this generator to generate adversarial examples for testing images and videos. The adversarial examples are then fed into object detectors to accomplish the attacking task.

\subsection{Network Architecture}
 Essentially, the adversarial example generation can be formulated into an image-to-image translation problem. The clean images or frames are input, and the adversarial images or frames are output. Therefore, we can refer to the training manner of pix2pix \cite{pix2pix2017}. In this paper, we utilize the network architecture in \cite{xiao2018generating} for ImageNet images, that is the first framework to generate adversarial examples using a pix2pix adversarial generative network. 
The generator is an encoder-decoder network with 19 components.  The discriminator is similar to ResNet-32 for CIFAR-10 and MNIST. Please refer to \cite{xiao2018generating} for the detailed structure of the generator and discriminator. 

\subsection{Loss Functions}
To simultaneously attack the current two kinds of object detectors, we need additional loss functions on the basis of Eq.(\ref{eq:eq1}) and Eq.(\ref{eq:eq2}).
To fool Faster-RCNN detector, DAG \cite{xie2017adversarial} uses a misclassify loss to make the predictions of all proposal regions go wrong. We  also integrate this loss. The class loss function is defined as follows:
%----------------------------Eq.3------------------------------------
\begin{equation}\label{eq:eq3}
\mathcal{L}_{DAG}(\mathcal{G})= \mathbb{E}_I[\sum_{n=1}^N[f_{l_n}(\textbf{X},t_n)-f_{\hat{l}_n}(\textbf{X},t_n)]],
\end{equation}
%--------------------------------------------------------------------
where \textbf{X} is the extracted feature map from the feature network of Faster-RCNN on $I$, and $\tau=\{t_1,t_2,...,t_N\}$ is the set of all proposal regions on \textbf{X}. $t_n$ is the $n$-th proposal region from the Region Proposal Network (RPN). $l_n$ is the ground-truth label of $t_n$, and $\hat{l}_n$ is  the wrong label randomly sampled from other incorrect classes. $f_{l_n}(\textbf{X}, t_n) \in \mathbb{R}^C$ denotes the classification score vector (before softmax normalization) on the $n$-th proposal region. In the experiments, we pick the proposals with score $\geq$0.7 to form $\tau=\{t_1,t_2,...,t_N\}$.  

DAG loss function is specially designed for attacking Faster-RCNN, therefore its transferability to other kinds of models is weak. To address this issue, we propose the following multi-scale attention feature loss:
%----------------------------Eq.4------------------------------------
\begin{equation}\label{eq:eq4}
\mathcal{L}_{Fea}(\mathcal{G})= \mathbb{E}_I[\sum_{m=1}^M\vert\vert \textbf{A}_m\circ(\textbf{X}_m-\textbf{R}_m)\vert\vert_2],
\end{equation}
%--------------------------------------------------------------------
where $\textbf{X}_m$ is the extracted feature map in the $m$-th layer of the feature network. $\textbf{R}_m$ is a randomly predefined feature map, and is fixed during training.  To fool detectors, only the regions of foreground objects need perturbing. We use the attention weight $\textbf{A}_m$ to measure the objects in $\textbf{X}_m$.  $\textbf{A}_m$ is computed based on the region proposals of RPN.  We let $s_n$ denote the  score of region proposal $t_n$. For each pixel in the original image, we collect all the region proposals covering this pixel, and compute the sum $\textbf{S}$ of these proposals' scores $s_n$, and then divide $\textbf{S}$ by the number of proposals $N$ to obtain the attention weight in the original image. Finally, $\textbf{A}_m$ is obtained by mapping the original attention weight to the $m$-th feature layer. For the pixels within objects, their weights will have large values and vice verse.  $\circ$ is the Hadamard product between two matrices.
By making $\textbf{X}_m$ as close as $\textbf{R}_m$, Eq.(\ref{eq:eq4}) enforces the attention feature maps to be random permutation, and thus manipulates the feature patterns of foreground objects.  $\textbf{R}_m$ can also be replaced by other feature maps different from $\textbf{X}_m$.
In the experiments, we choose the Relu layer after conv3-3 and the Relu layer after conv4-2 in VGG16 to manipulate their feature maps. To compute $\textbf{A}_m$, we use the top 300 region proposals according to their scores.

Finally, our full objective can be expressed as:
%----------------------------Eq.4------------------------------------
\begin{equation}\label{eq:eq5}
\mathcal{L}=\mathcal{L}_{cGAN}+\alpha\mathcal{L}_{L_2}+\beta\mathcal{L}_{DAG}+\epsilon\mathcal{L}_{Fea},
\end{equation}
%--------------------------------------------------------------------
where $\alpha,\beta,\epsilon$ are the relative importance of each objective. We set $\alpha=0.05, \beta=1$. For $\epsilon$, we set $1 \times10^{-4}$ and $2 \times10^{-4}$ for the selected two layers, respectively. 
 $\mathcal{G}$ and $\mathcal{D}$ are obtained
by solving the minmax game $arg min_\mathcal{G} max_\mathcal{D} \mathcal{L}$.
To optimize our networks under Eq.(\ref{eq:eq5}), we follow the standard approach
from \cite{pix2pix2017} and
apply the Adam solver \cite{kingma2014adam}. The best weights are obtained after 6 epochs. 

\section{Experiments}
\subsection{Datasets}
\textbf{Image Detection}
In order to train the adversarial generator in UEA, we use the training dataset of PASCAL VOC 2007 with totally 5011 images. They are categorized into 20 classes.  In testing phase,  we use the PASCAL VOC 2007 testing set \cite{pascal-voc-2007} with 4952 images. 

\textbf{Video Detection}
For video detection, we use ImageNet VID dataset\footnote{http://bvisionweb1.cs.unc.edu/ILSVRC2017/download-videos-1p39.php}. There are 759 video snippets for training set, and 138 for testing set. The frame rate is 25 or 30 fps for most snippets. There are 30 object categories, which are a subset of the categories in the ImageNet dataset. 
\subsection{Metrics}
We use two metrics: attacking performance against object detectors;  the generating time for adversarial examples. 

 \textbf{Fooling Rate}: to test the fooling rate of different attacking methods, we use the mAP drop (mean Average Precision). The mAP is usually to evaluate the recognition accuracy of object detectors both for image and video data. If the adversary is strong, detectors will achieve a lower mAP on adversarial examples than clean examples. The reducing error can be used to measure the attacking methods.

 \textbf{Time}: to tackle with video data, the time for generating adversarial examples is important. In the experiments, we report the processing time for each image (frame) against different attacking methods.

\subsection{Threat Models}
\textbf{Image Detection} In this setting, our goal is to simultaneously attack the proposal based detectors and regression based detectors. We select two representative methods: 
Faster-RCNN and SSD300. There are a lot of implementation codes for them. Here we use the Simple Faster-RCNN\footnote{https://github.com/chenyuntc/simple-faster-rcnn-pytorch}
and torchCV SSD300\footnote{https://github.com/kuangliu/torchcv}. We retrain their models on PASCAL VOC training datasets. Specifically, Faster-RCNN is trained on the PASCAL VOC 2007 training dataset, and tested  on the PASCAL VOC 2007 testing set. The detection accuracy (mAP) reaches 0.70. SSD300 is trained on the hybrid dataset consisting of PASCAL VOC 2007 and 2012 training set, and tested on the PASCAL VOC 2007 testing set. The mAP reaches 0.68.

\textbf{Video Detection} The current video detection methods are based on image detection. They usually perform image detection on  key frames, and then propagate the results to other frames \cite{zhu2017flow,zhu2017deep,chen2018optimizing}. However, as shown in \cite{zhu2017deep}, the detection accuracy of these efficient methods cannot even outperform the simple dense detection method, that densely runs the image detection on each frame in a video. In \cite{zhu2017flow}, although their method beats dense detection method, they cost more time. If they reduce the processing time, the accuracy also falls below the dense detection.  Therefore, we choose the dense detection method as the threat model. We argue if the dense detection method is successfully attacked, the efficient methods will also fail. 

\subsection{Results on Image Detection}
\subsubsection{Comparisons with  State-of-the-art Methods}
The current state-of-the-art attacking method for image detection is DAG. Therefore, we use DAG as our compared method. For that, we generate adversarial image using DAG and UEA, respectively, and then perform the same Faster-RCNN and SSD300 on the adversarial examples to observe the accuracy drop (compared with the accuracy on clean images). Meanwhile, we also check  the generating time (Time). The comparison results are reported in Table \ref{tab:tab2}.

\begin{table}[htbp]
\vspace{-.2cm}
\caption{The comparison results between DAG and UEA. } \centering
\vspace{-.2cm}
\begin{tabular}{|p{2cm}<{\centering}|p{2cm}<{\centering}|p{1.4cm}<{\centering}|p{1.4cm}<{\centering}|}
\hline
 \multirow{2}{*}{\textbf{Methods} } & \multicolumn{2}{c|}{Accuracy (mAP)} &\multicolumn{1}{c|}{\multirow{2}{*}{Time (s)}} \\
 \cline{2-3}
 & Faster-RCNN & SSD300 & \multicolumn{1}{c|}{} \\
 \hline
 Clean Images &0.70  &0.68   &\verb|\|   \\
 \hline
 DAG &0.05  &0.64  & 9.3  \\
 \hline
 UEA &\textbf{0.05}  &\textbf{0.20}  &\textbf{0.01}  \\
 \hline
\end{tabular}
\label{tab:tab2}
\vspace{-.2cm}
\end{table}

From the table, we see: \textbf{(1):} Both DAG and UEA work well on attacking Faster-RCNN detector. They achieve the same 0.65 accuracy drop (0.70-0.05). This is expected because DAG and UEA formulate the same class loss of Eq.(\ref{eq:eq3}) into their methods, and they perform the white-box attack against Faster-RCNN. \textbf{(2):}  DAG cannot attack SSD detector, the accuracy drop is only 0.04 (0.68-0.64). By contrast, UEA obtains a 0.48 accuracy drop (0.68-0.20), which is 12 times larger than DAG. This verifies the weak black-box attacking ability of DAG. Instead, UEA integrates a feature loss to manipulate the shared feature networks between Faster-RCNN and SSD. The feature loss enhances the transferability and black-box attacking ability to other kinds of detectors.  Theoretically, UEA is able to attack a large class of object detectors besides SSD and Faster-RCNN, because the majority of object detectors use the feature network. \textbf{(3):}  As for the generating time of adversarial examples, UEA is almost 1000 times faster than DAG (0.01 vs 9.3). The efficiency is helpful to tackle with video data. Even for the video with 100 frames, UEA will only cost one second to pollute all the frames. 

\begin{figure}[t]
\centering\includegraphics[width=0.4\textwidth]{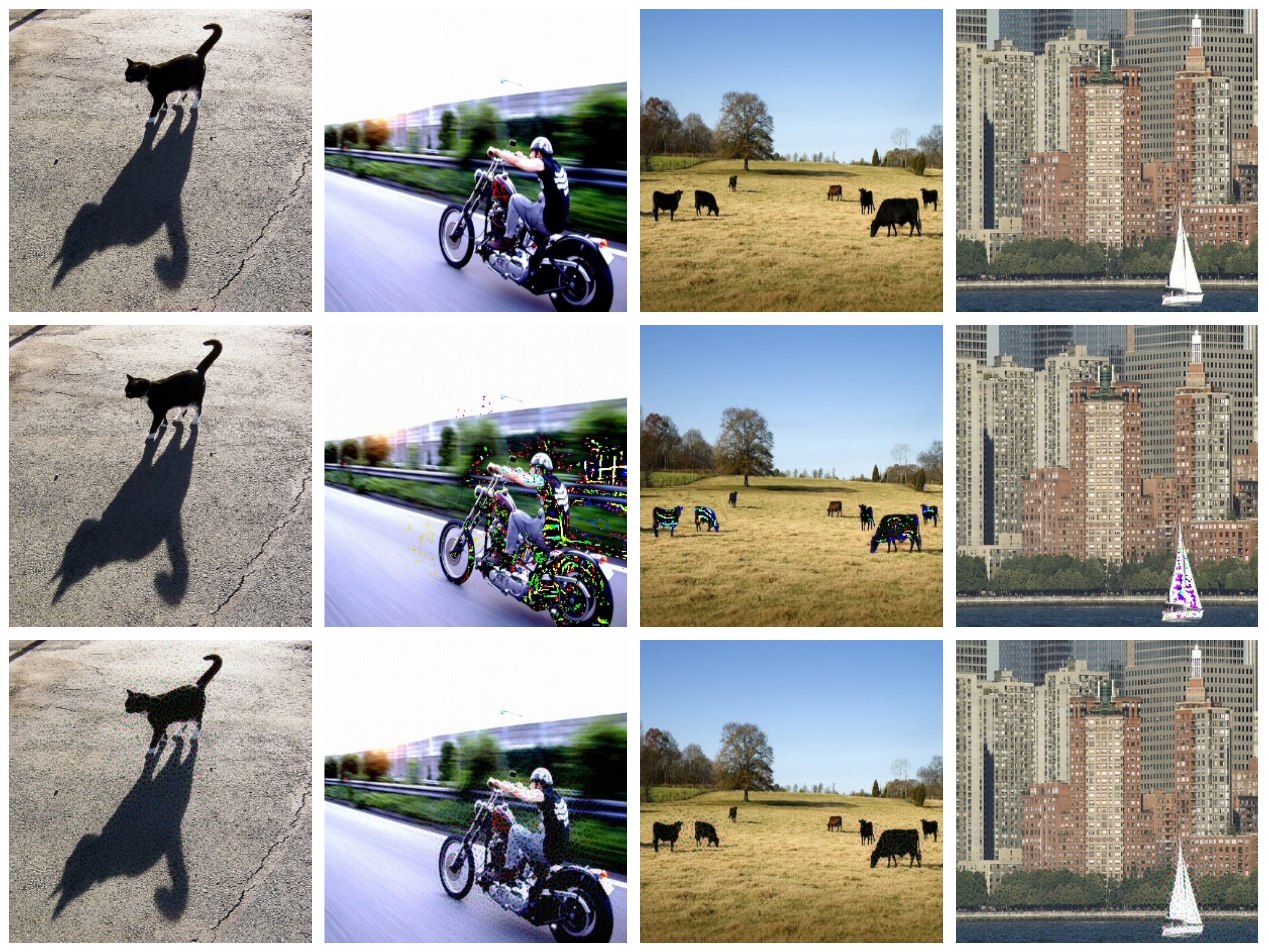}
\vspace{-.2cm}
\caption{The perceptibility comparison of adversarial images. The first row is clean images. The second row is output by DAG (the iteration is 1, 81, 133, 41, respectively). The third row is our output.  }
\label{fig:figure9}
\vspace{-.4cm}
\end{figure}

We also evaluate the perceptibility of adversarial examples. Figure \ref{fig:figure9} gives the comparisons.  As an optimization method, DAG is highly relevant with different images. Their perturbations are increasing with the rising iterations. For example, DAG only costs 1 iteration for ``cat" image, and the perturbations are imperceptible. But for  ``motorbike" image, DAG costs 81 iterations, and the perturbations are very obvious. The ``cow" and ``boat" images have the similar trend.  UEA is a generative method. We see the adversarial examples are always imperceptible, and almost the same as the clean images. 

\subsubsection{Ablation Study of UEA}
\begin{figure}[t]
\centering\includegraphics[width=0.23\textwidth]{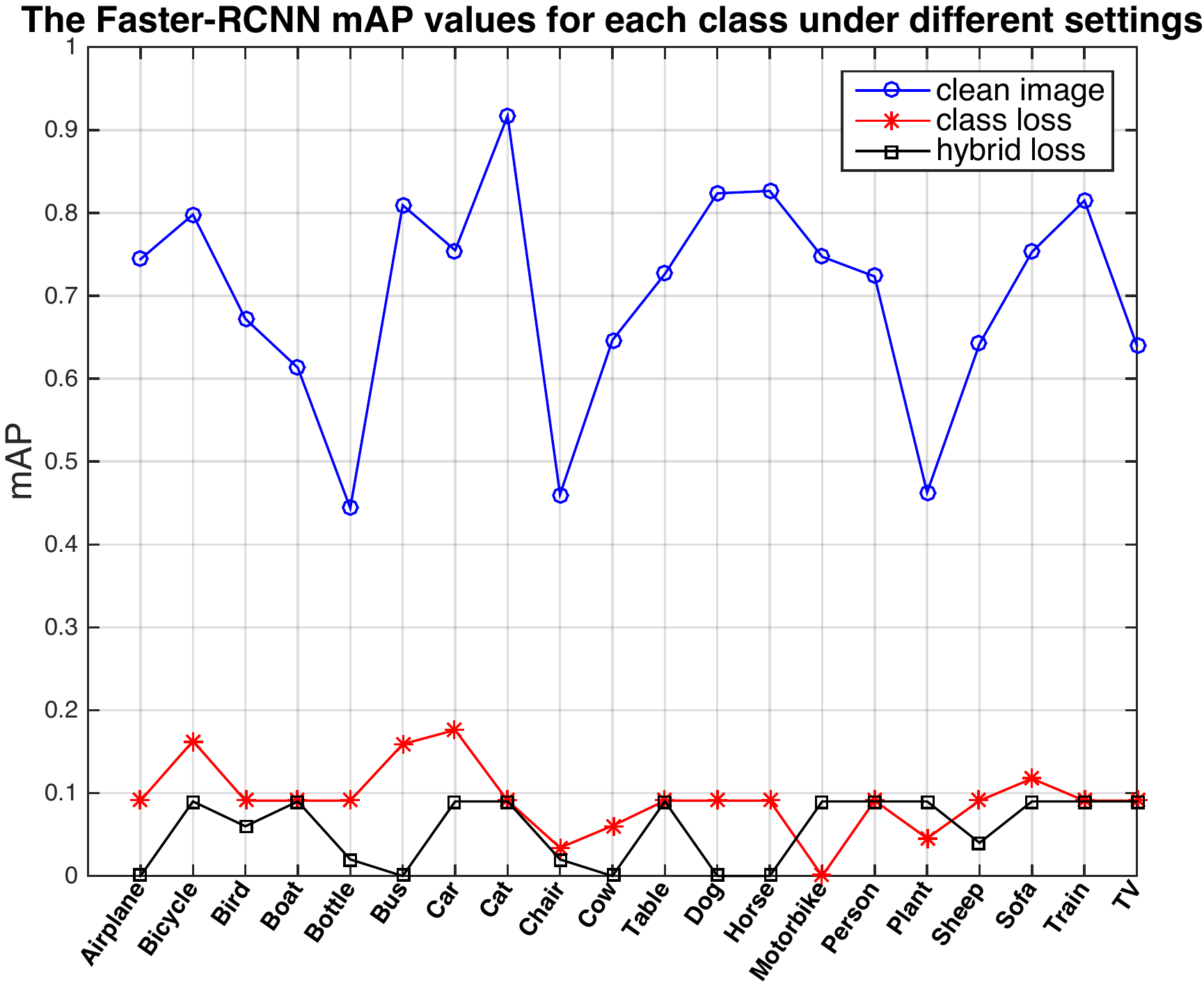}
\centering\includegraphics[width=0.22\textwidth]{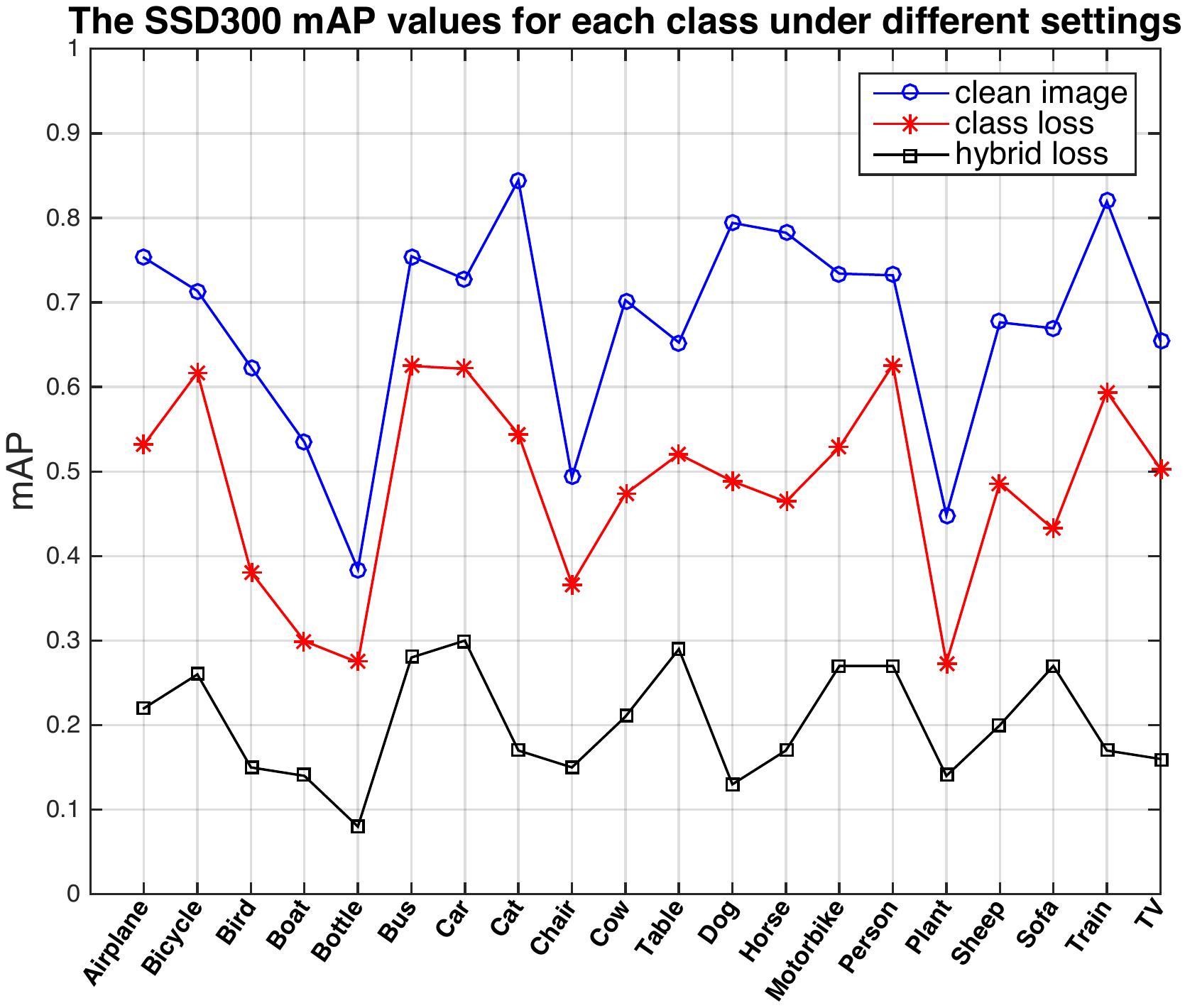}
\vspace{-.4cm}
\caption{The ablation study of UEA  for each category detection.}
\label{fig:figure3}
\vspace{-.2cm}
\end{figure}

Now we look into the ablation study of UEA. As introduced in Section 3, UEA utilizes two key loss functions in the training phase. The first is class loss, i.e., Eq.(\ref{eq:eq3}), and the second is multi-scale attention feature loss, i.e., Eq.(\ref{eq:eq4}). We study the function of each loss, and report the results for each category detection in Figure \ref{fig:figure3}. In this figure, ``class loss" denotes the absence of ``feature loss" in UEA, i.e., only the DAG's loss.  ``hybrid loss" is the full version of UEA with both ``class loss" and ``feature loss". From the figure, we see that ``class loss" works well on Faster-RCNN, but shows the limited attacking ability on SSD300. After adding the proposed ``feature loss", UEA  has the similar attacking performance with ``class loss" on Faster-RCNN, but shows stronger attacking ability on SSD300.  These results demonstrate that hybrid the high-level class loss and low-level feature loss is a reasonable choice for attacking object detectors.

\begin{figure}[t]
\centering\includegraphics[width=0.45\textwidth]{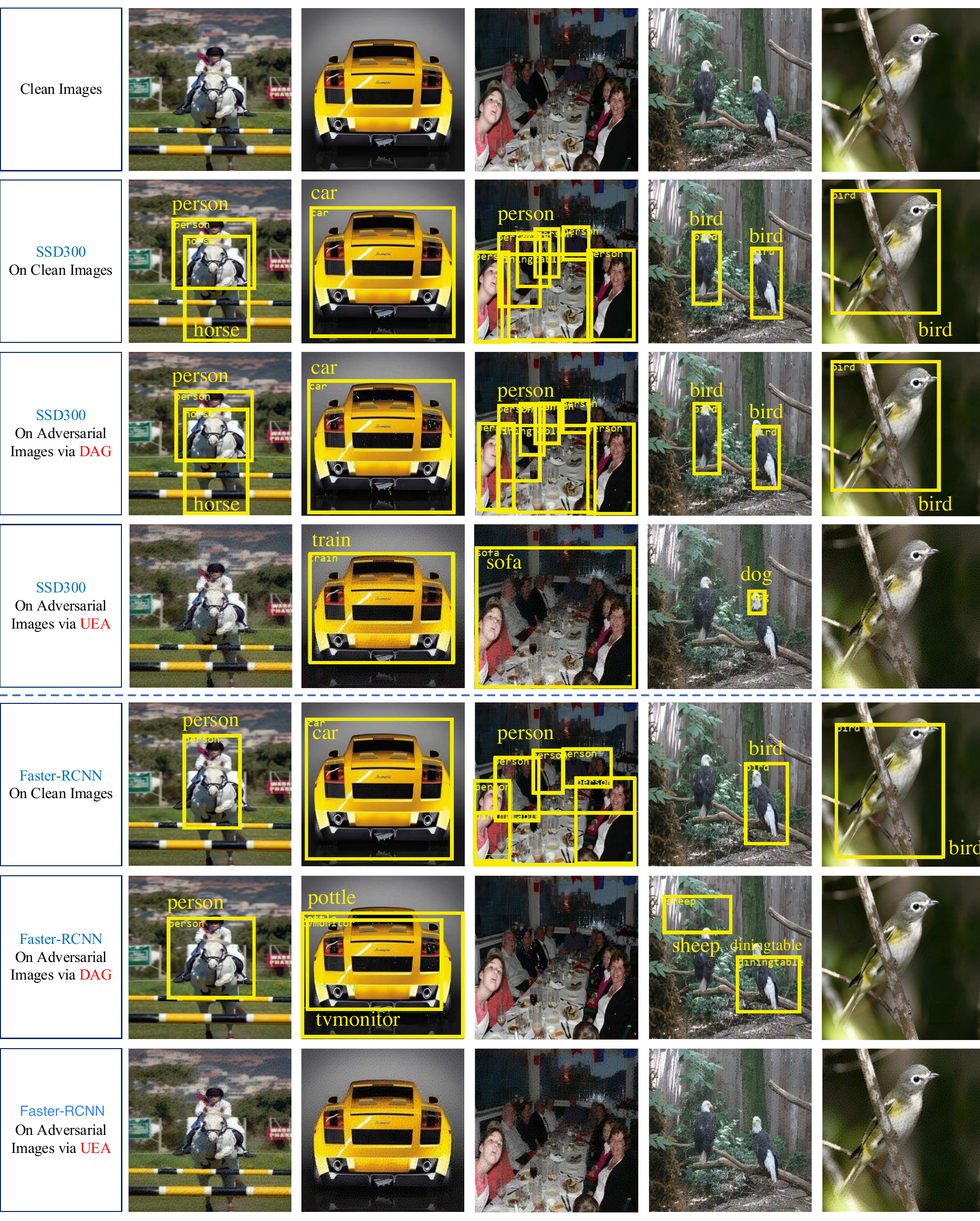}
\vspace{-.2cm}
\caption{The qualitative comparisons between DAG and UEA versus Faster-RCNN and SSD300. Please see the texts for details.}
\label{fig:figure4}
\vspace{-.4cm}
\end{figure}

\subsubsection{Qualitative Comparisons}
We give some qualitative comparisons between DAG and UEA in Figure \ref{fig:figure4}. From the figure, we see both Faster-RCNN and SSD300 work well on the clean images, and detect the correct bounding boxes and labels. For DAG, it succeeds to attack Faster-RCNN (see the sixth row where Faster-RCNN doesn't detect any object on two images and predicts wrong labels on three images). However, SSD300 still works well on the adversarial examples generated by DAG. For UEA, Faster-RCNN cannot detect any bounding box on five adversarial examples, and SSD300 detects wrong objects on two images and  zero detection on three images. 

\begin{figure}[t]
\vspace{-.2cm}
\centering\includegraphics[width=0.45\textwidth]{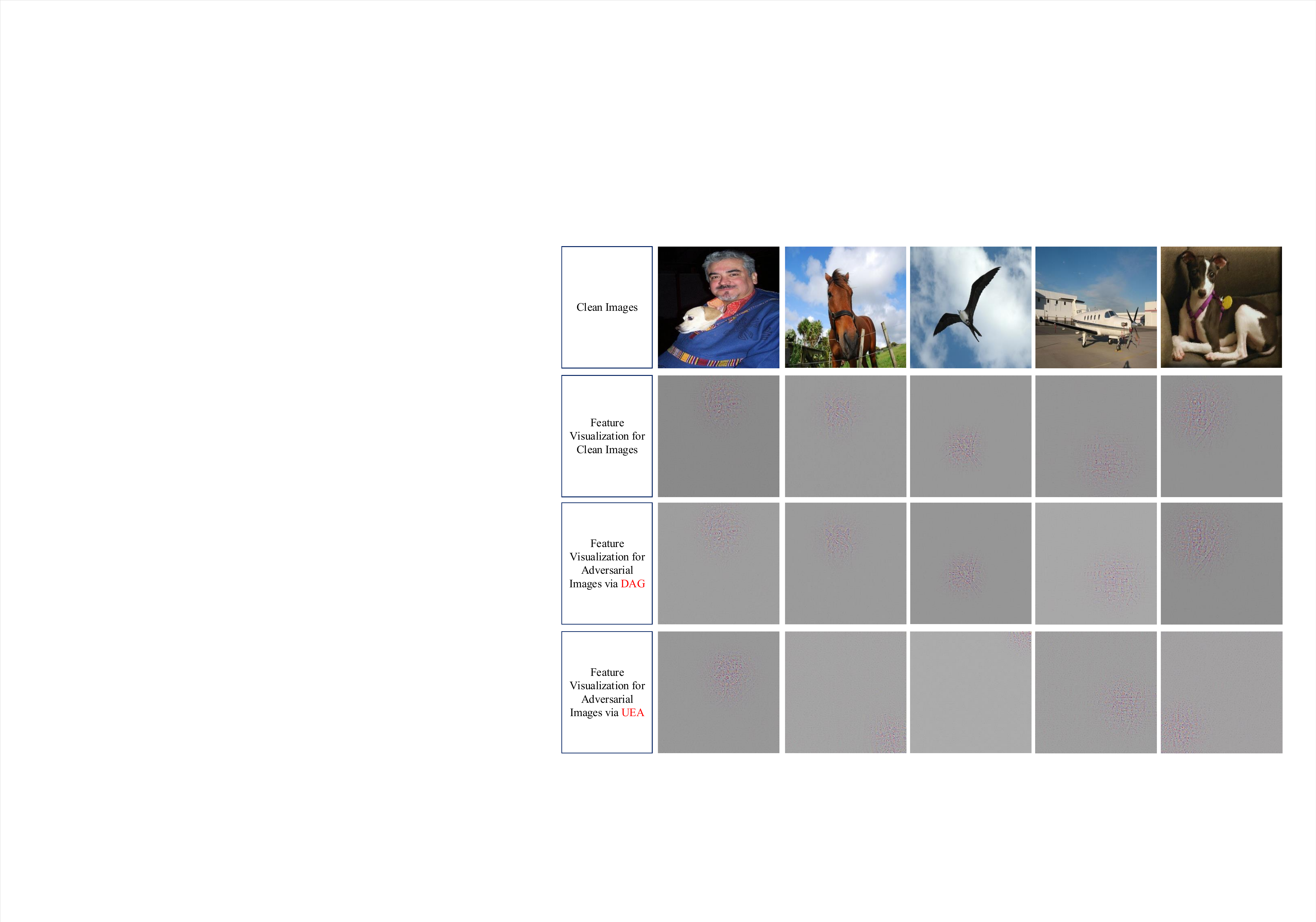}
\vspace{-.2cm}
\caption{The feature visualization of adversarial examples via DAG and UEA, respectively. Please see the texts for details. }
\label{fig:figure5}
\vspace{-.2cm}
\end{figure}

To better show the intrinsic mechanism of UEA, we visualize the feature maps extracted from adversarial examples via DAG and UEA, respectively. Because Faster-RCNN and SSD300 utilize the same VGG16 as their feature network, we select the feature maps extracted on conv4 layer and visualize them using the method in \cite{zeiler2014visualizing}. From Figure \ref{fig:figure5}, we see that the feature maps via UEA have been manipulated. Therefore, the Region Proposal Network within Faster-RCNN cannot output the available proposal regions, and thus Faster-RCNN doesn't detect any bounding box. For SSD300, the manipulated features make the regression operation not work, leading to wrong or vacant predictions. 

\begin{figure}[t]
\centering\includegraphics[width=0.45\textwidth]{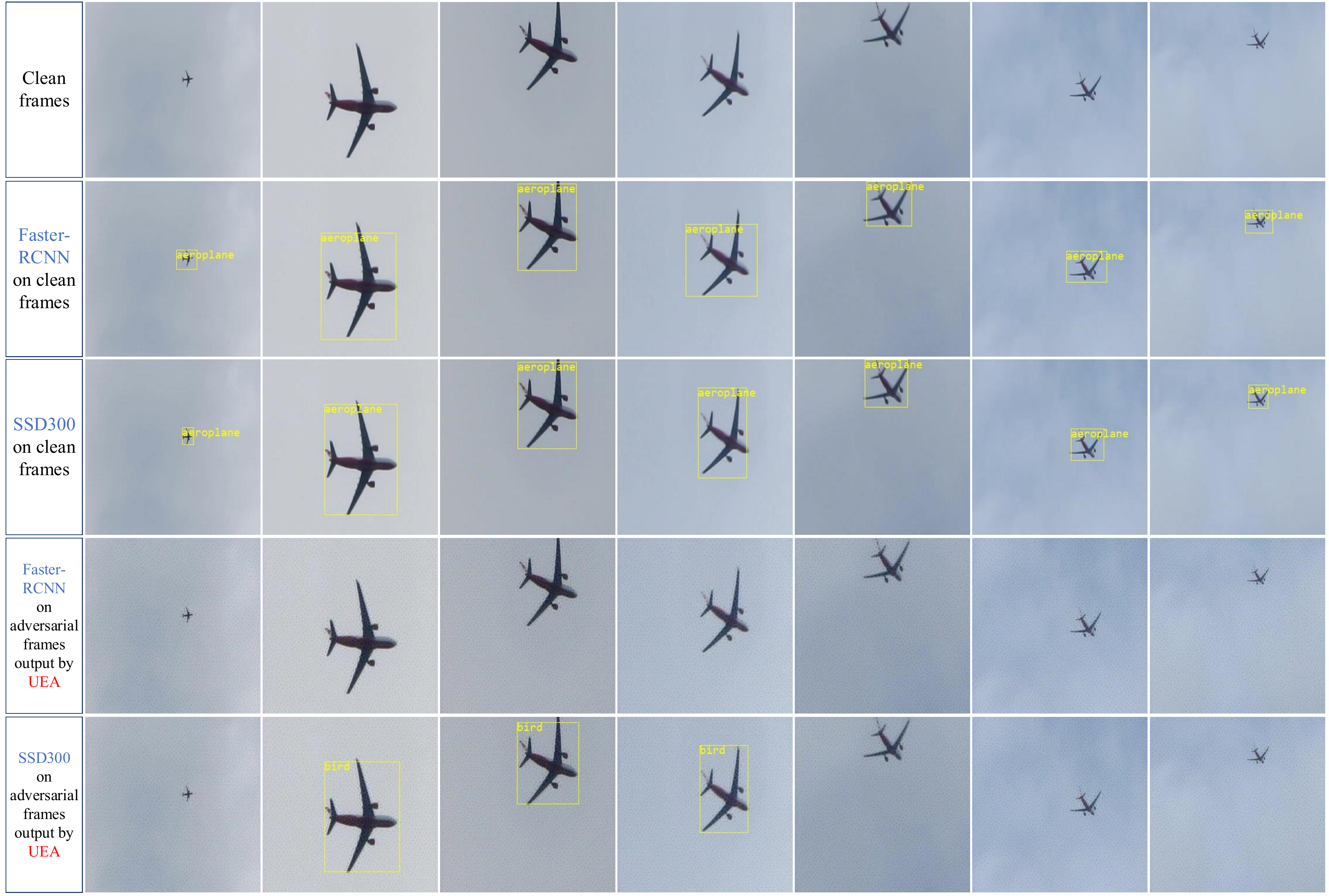}
\caption{The qualitative result of attacking  video object detection }
\label{fig:figure8}
\vspace{-.4cm}
\end{figure}

\begin{figure}[t]
\centering\includegraphics[width=0.23\textwidth]{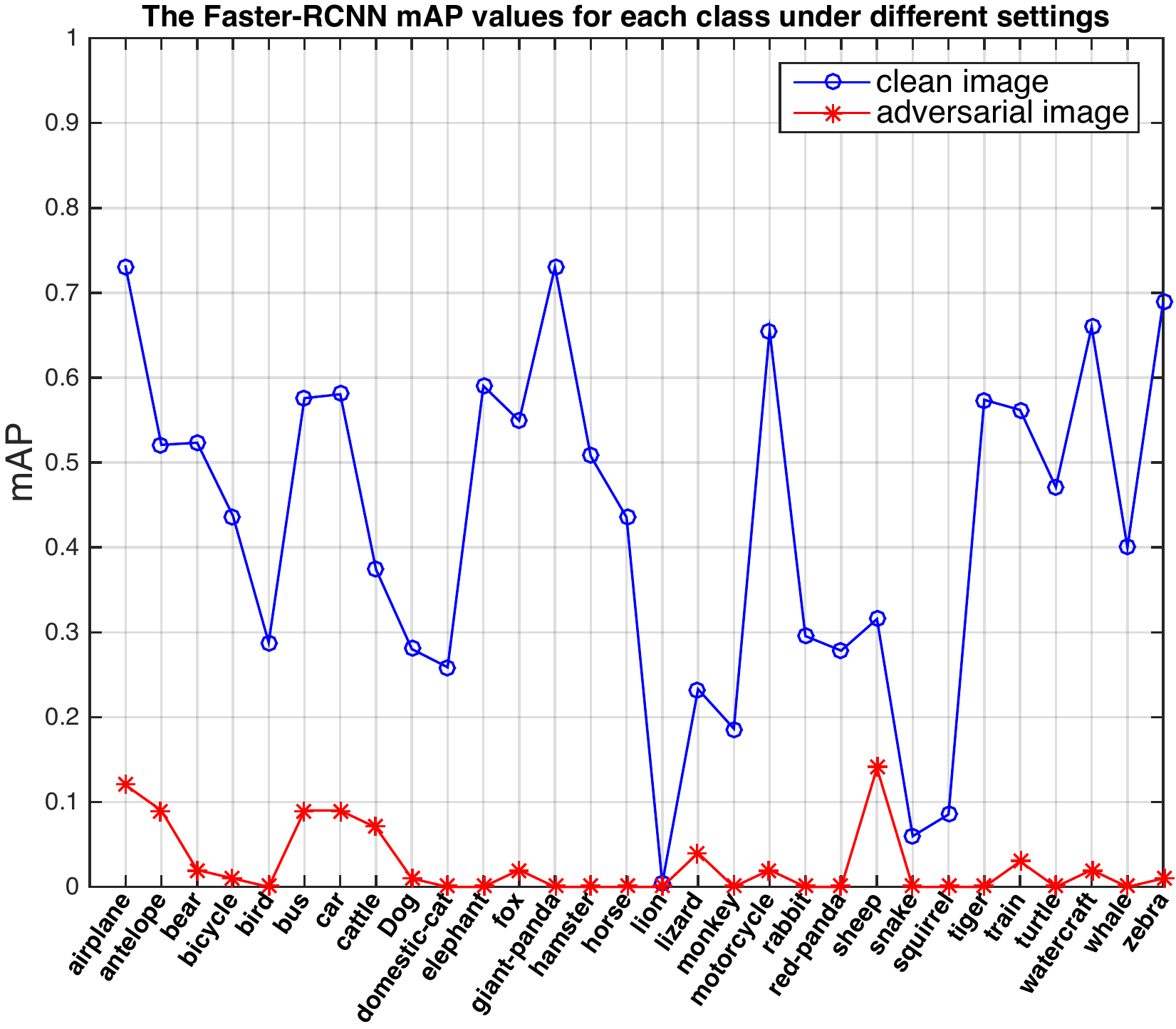}
\centering\includegraphics[width=0.23\textwidth]{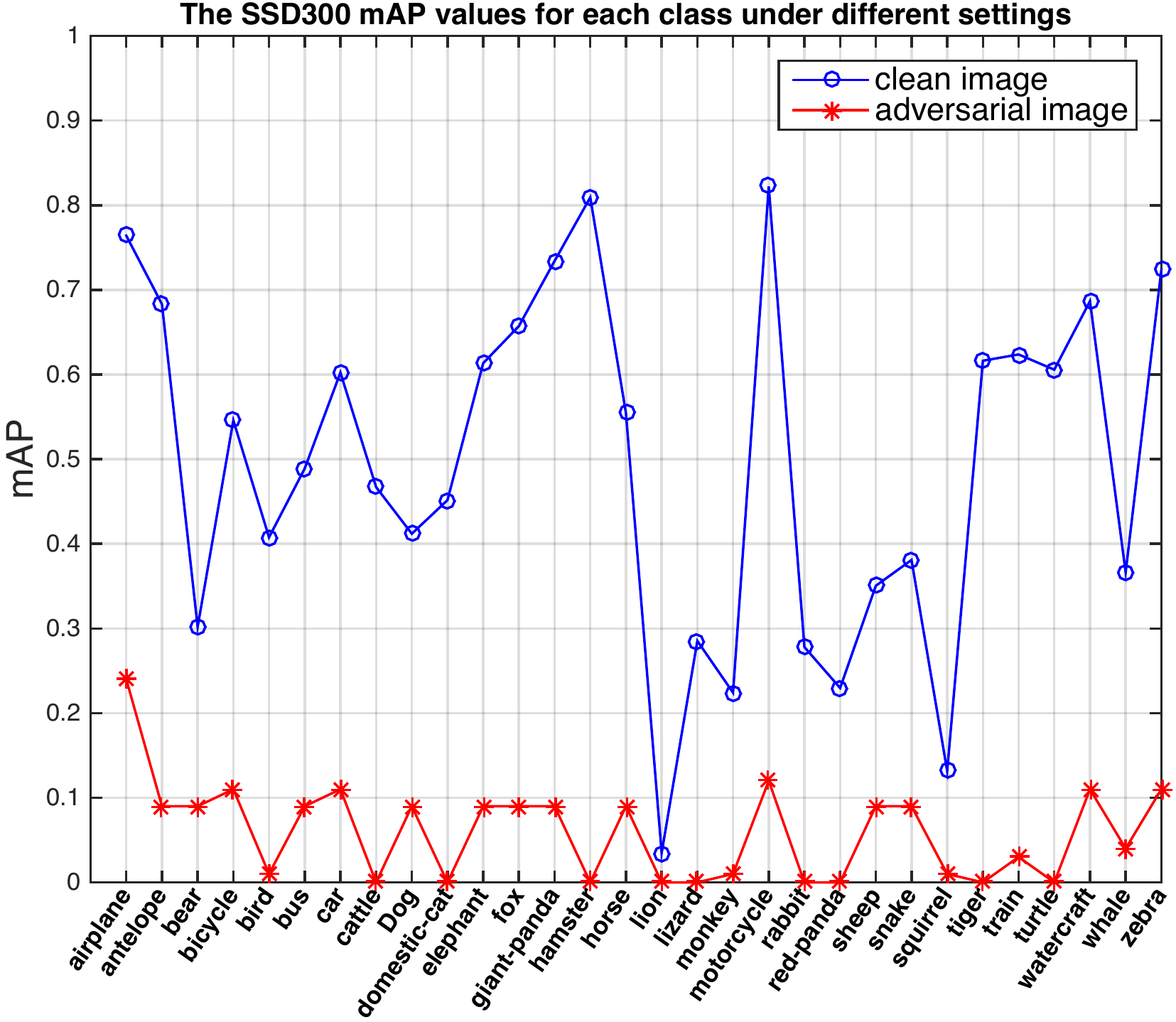}
\vspace{-.2cm}
\caption{The detecting performance of object detectors on clean videos and adversarial videos for each category detection.}
\label{fig:figure10}
\vspace{-.4cm}
\end{figure}

\subsection{Results on Video Detection}
In this section, we report the results on video object detection. We here use the ImageNet VID dataset. 
As discussed in section 4.3, we attack the dense frame detection methods. Specifically, we train Faster-RCNN and SSD300 on ImageNet VID dataset, and then run the detectors on each frame in the testing video. We believe that if the dense frame detection method can be successfully attacked, other efficient methods will be also fooled. Figure \ref{fig:figure8} reports the qualitative attacking results versus ``airplane". Similarly, UEA simultaneously fools Faster-RCNN and SSD300.  
In Figure \ref{fig:figure10}, we list the comparison mAP performance for each category detection. Readers can have a clear observation for the mAP drop for each class on ImageNet VID. 

Table \ref{tab:tab5} shows the quantitative attacking performance of UEA on ImageNet VID. 
Specifically, we train Faster-RCNN and SSD300 on the training set of ImageNet VID, and run the trained detectors on the testing set. In addition, we use UEA to generate the corresponding adversarial videos for the testing set of ImageNet VID, and then run the same detectors.   In Table  \ref{tab:tab5}, we see UEA achieves 0.40 mAP drop for Faster-RCNN, and 0.44 mAP drop for SSD300, which shows UEA achieves a good attacking performance in the video data. 

We here use the VGG16 based Faster-RCNN and SSD300. \cite{zhu2017deep} shows that if we use ResNet 101 as the backbone network, and replace Faster-RCNN with FCN \cite{dai2016r} as the object detector, the original mAP will reach 0.73. 
Because our paper aims at measuring the attacking ability of UEA, rather than the detecting performance, the mAP drop is the key metric, rather than mAP.  Therefore, we here don't  use ResNet 101+FCN. Similarly, we also don't use the SSD500, although it has better detection than SSD300. The current mAP drop has verified the powerful attacking ability of UEA both against the proposal based detector (Faster-RCNN) and regression based detector (SSD300).  We believe that if we use the advanced object detectors, the mAP drop will also improve. 

\begin{table}[h]
\vspace{-.3cm}
\caption{The attacking performance of UEA on video detection. } \centering
\vspace{-.2cm}
\begin{tabular}{|p{2cm}<{\centering}|p{2cm}<{\centering}|p{1.4cm}<{\centering}|p{1.4cm}<{\centering}|}
\hline
 \multirow{2}{*}{\textbf{Methods} } & \multicolumn{2}{c|}{Accuracy (mAP)} &\multicolumn{1}{c|}{\multirow{2}{*}{Time (s)}} \\
 \cline{2-3}
 & Faster-RCNN & SSD300 & \multicolumn{1}{c|}{} \\
 \hline
Clean Videos   &0.43  &0.50  &\verb|\| \\
\hline
UEA   &0.03  &0.06    &0.3s\\
\hline
mAP drop &\textbf{0.40} &\textbf{0.44} &\verb|\|  \\
\hline
\end{tabular}\label{tab:tab5}
\vspace{-.6cm}
\end{table}

\section{Conclusion}
In this paper, we proposed the Unified and Efficient Adversary (UEA). UEA was able to efficiently generate adversarial examples, and its processing time was 1000 times faster than the current attacking methods. Therefore, UEA could deal with not only image data, but also video data. More importantly, UEA had better transferability than the existing attacking methods, and thus, it could meanwhile attack the current two kinds of representative object detectors. Experiments conducted on PASCAL VOC and ImageNet VID verified the effectiveness and efficiency of UEA.

{\small
\bibliographystyle{named}
\bibliography{egbib}
}

\end{document}